\PassOptionsToPackage{table}{xcolor}
\documentclass[sigconf,screen,nonacm]{acmart}
\AtBeginDocument{%
  }

\usepackage{multirow}
\usepackage{algorithm}
\usepackage{algorithmic}

\begin{document}

\title{Mitigating Visual Degradation in MLLMs via Spatial-Spectral Visual Anchor Learning}

\author{Qianlong Yang}
\authornote{This work was conducted during an internship at Great Wall Motor.}
\affiliation{%
  \institution{China University of Petroleum \\ (East China)}
  \city{Qingdao}
  \state{Shandong}
  \country{China}
}

\email{z24090014@s.upc.edu.cn}

\author{Bowen Ye}
\affiliation{%
  \institution{Shanghai Jiao Tong University}
  \city{Shanghai}
  \country{China}
}
\email{yebowen1025@sjtu.edu.cn}
  
\author{Xianda Guo}
\correspondingauthor
\affiliation{%
  \institution{School of Computer Science, Wuhan University}
  \city{Wuhan}
  \country{China}
}
\email{xianda\_guo@163.com}

\author{Yanlun Peng}
\affiliation{%
 \institution{Great Wall Motor}
 \city{Baoding}
 \country{China}}
 \email{yanlunpeng@gwm.cn}

\author{Wenke Huang}
\affiliation{%
  \institution{Nanyang Technological University}
  \city{Singapore}
  \country{Singapore}}
  \email{wenke.huang@ntu.edu.sg}

\author{Hongyuan Zhang}
\affiliation{%
  \institution{The University of Hong Kong}
  \city{Hong Kong}
  \country{China}}
\email{hyzhang98@gmail.com}

\author{Yulei Jia}
\correspondingauthor
\affiliation{%
  \institution{China University of Petroleum \\ (East China)}
  \city{Qingdao}
  \state{Shandong}
  \country{China}}
  \email{jiayl@upc.edu.cn}

\renewcommand{\shortauthors}{Qianlong Yang et al.}

\begin{abstract}
  Despite the progress of multimodal large language models (MLLMs), they continue to exhibit deficiencies in visual perception. Following visual instruction tuning, internal MLLM representations rapidly deviate from their original semantic states during inference, causing severe information degradation. While existing methods attempt to leverage external vision foundation models (VFMs) to align internal representations, we find that direct alignment with VFMs enhances visual semantics but fails to mitigate representation deviation. To address this, we propose Spatial-Spectral Visual Anchor Learning (SSVAL). The core of SSVAL is Visual Anchor Prompt Injection (VAPI), which introduces prompts that absorb rich knowledge from external VFMs during training, enabling them to serve as stable visual anchors that mitigate representation deviation during inference. Additionally, we incorporate auxiliary spatial and frequency-domain representation alignment losses to provide complementary vision-specific supervision at intermediate LLM layers. Extensive experiments demonstrate that SSVAL significantly outperforms existing methods. Code are available on our \href{https://msls38.github.io/SSVAL/}{project page}.
\end{abstract}




\maketitle

\section{Introduction}
\begin{figure*}[t]
  \centering
  \includegraphics[width=\linewidth]{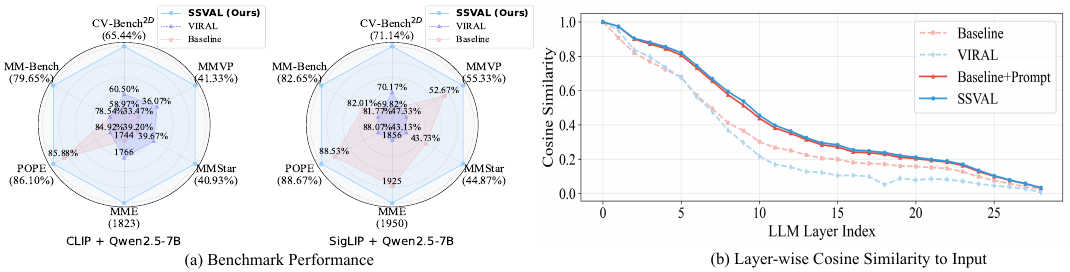}
  \Description{Bar chart and line chart showing SSVAL performance on multimodal benchmarks and layer-wise similarity analysis compared to Baseline and VIRAL.}
  \caption{Analysis of SSVAL effectiveness and performance comparison. (a) SSVAL demonstrates superior comprehensive performance across various multimodal benchmarks. (b) Layer-wise similarity analysis shows that SSVAL (ours) better preserves visual representations in deeper layers compared to conventional visual instruction tuning (Baseline)~\cite{liu2023visual} and Visual Representation Alignment (VIRAL)~\cite{yoon2025visual}.}
  \label{fig:similarity}
\end{figure*}

Recent advancements in MLLMs~\cite{dai2023instructblip,qiao2025class,chen2024internvl,jiang2025mitigating,liu2023visual,bai2025qwen2,qiao2025bidirectional,MuNG,ViewMask,UNIFIER,yin2026holo,guo2026surds,zhang2024InstructLanguage} focus on the critical challenge of effectively aligning visual and linguistic modalities. Visual instruction tuning~\cite{liu2023visual} addresses this by employing a lightweight projector to bridge vision encoders~\cite{radford2021learning,oquab2023dinov2,ravisam,sun2023eva} with pretrained LLMs~\cite{touvron2023llama,achiam2023gpt,yang2025qwen3,team2025kimi}. This projector functions as a semantic adapter, mapping visual representations into the LLM's input embedding space. This integration empowers LLMs to comprehend visual tokens as if they were a foreign language, thereby facilitating profound cross-modal reasoning and achieving substantial performance improvements across a wide range of downstream tasks.

Despite the success of MLLMs, they still suffer from significant visual perception deficiencies~\cite{tong2024eyes,kar2024brave,yuksekgonuland}. A natural question arises: \textit{why do MLLMs, equipped with powerful vision encoders, still struggle with visual perception?} While prior works attribute this to inadequate vision encoders~\cite{li2024llava,chen2024internvl,karamcheti2024prismatic} or projectors~\cite{cha2024honeybee,li2024llava-next,lin2023sphinx}, they often overlook a critical yet under-explored phenomenon occurring \textit{inside} the LLM itself. In this work, we identify this phenomenon as \textbf{visual representation deviation}: after visual instruction tuning, the internal visual representations within LLM layers progressively deviate from their original semantic states during inference, leading to severe information degradation in deeper layers~\cite{fu2025hidden}. To quantify this phenomenon, we measure the cosine similarity between the input visual representations and intermediate-layer representations across LLM depths. As shown in Fig.~\ref{fig:similarity}~(b), the similarity drops sharply as the layer depth increases, confirming that visual information is progressively lost during forward propagation. This degradation directly explains the visual perception failures observed in downstream tasks.

An intuitive solution to this problem is to directly align intermediate LLM representations with external Vision Foundation Models (VFMs)~\cite{oquab2023dinov2,simeoni2025dinov3,lin2025depth}, which provide rich and stable visual features as alignment targets. However, we reveal a counter-intuitive finding: as shown in Fig.~\ref{fig:similarity}~(b), direct VFM alignment primarily enhances the quality of visual representations but \textit{fails to mitigate the deviation itself}. The representations still diverge at a similar rate across layers, suggesting that simply injecting stronger visual features cannot anchor the representations against progressive drift. This observation reveals a fundamental limitation of direct alignment approaches~\cite{yoon2025visual}: they improve \textit{what} the representations encode, but not \textit{how stably} they are maintained across layers.

\begin{figure*}[t]
  \centering
  \includegraphics[width=\linewidth]{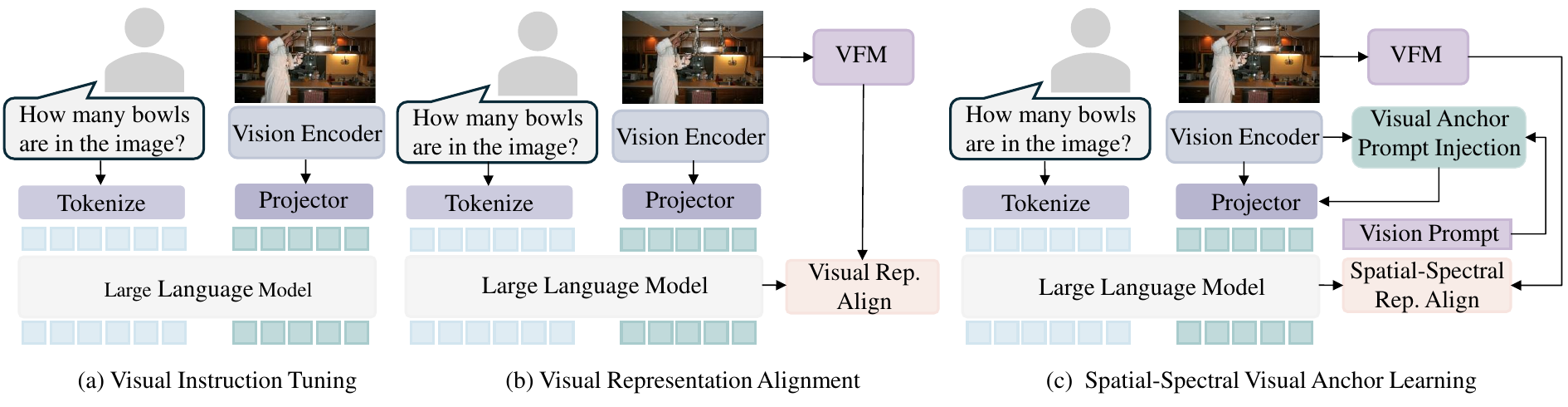}
  \caption{Comparison of MLLM architectures. (a) Visual Instruction Tuning~\cite{liu2023visual}. (b) Visual Representation Alignment~\cite{yoon2025visual}: aligns visual features with VFM. (c) SSVAL (Ours): introduces visual anchor prompts with auxiliary representation alignment.}
  \Description{Three architecture diagrams comparing different MLLM approaches: standard visual instruction tuning with a projector, visual representation alignment that aligns features with VFM, and SSVAL which introduces visual anchor prompts combined with spatial-spectral representation alignment.}
  \label{fig:intro}
\end{figure*}

This finding motivates our key insight: rather than directly modifying the feature distribution, we need a persistent \textbf{anchor mechanism} that absorbs VFM knowledge during training and continuously provides stable visual references to arrest the deviation. Based on this insight, we propose Spatial-Spectral Visual Anchor Learning (SSVAL), a framework designed to enhance the visual perception capabilities of MLLMs (see Fig.~\ref{fig:intro}). The core of SSVAL is \textbf{Visual Anchor Prompt Injection (VAPI)}, which introduces prompts that absorb rich visual knowledge from external VFMs during training, enabling them to serve as \textbf{visual anchors that continuously provide stable visual references} for each layer during inference. To complement VAPI, we further incorporate auxiliary representation alignment losses from multi-scale spatial and frequency-domain perspectives (\textbf{SpaRA} and \textbf{SpeRA}), which provide additional vision-specific supervision at intermediate LLM layers during training.

The main contributions of this paper are summarized as follows:
\begin{itemize}
    \item We identify and quantify the visual representation deviation phenomenon in MLLMs, and reveal a counter-intuitive finding: direct alignment with external VFMs enhances visual representations but fails to mitigate the deviation.
    \item We propose Visual Anchor Prompt Injection (VAPI), a mechanism that utilizes prompts as visual anchors to absorb external VFM knowledge during training and provide stable visual references, effectively alleviating representation deviation.
    \item Extensive experiments on multiple multimodal benchmarks demonstrate that SSVAL significantly outperforms existing methods (see Fig.~\ref{fig:similarity} (a)). We further show that auxiliary spatial and frequency-domain alignment losses provide complementary benefits during training.
\end{itemize}

\section{Related Works}

\subsection{Visual Prompt learning.}

Early research~\cite{jia2022visual,khattak2023maple,khattak2023self,han20232vpt,yu2025introducing} in Visual Prompt Learning was dedicated to efficiently adapting large-scale pre-trained models, such as CLIP~\cite{jiang2023clip}, to downstream tasks. By optimizing soft prompts via backpropagation through frozen backbones, these methods have demonstrated superior generalization capabilities in few-shot scenarios. However, these have primarily focused on exploring efficient prompt formats under data-scarce conditions. Subsequently, PromptKD~\cite{li2024promptkd} innovatively proposed treating prompts as 'knowledge distillers,' facilitating knowledge transfer from a teacher to a student model. Nevertheless, it remains limited to knowledge transfer between homogeneous models (i.e., CLIP to CLIP), failing to effectively absorb complementary knowledge—such as fine-grained geometric features from other vision foundation models. Consequently, it becomes essential to explore how visual prompts can serve as anchors to absorb heterogeneous VFM knowledge and mitigate visual representation deviation in MLLMs.

\subsection{Improving visual information degradation in MLLMs.}
Recent research indicates that visual information progressively degrades as it propagates through MLLM layers. Existing strategies primarily target input-stage enhancements—such as stronger vision encoders~\cite{kar2024brave,lu2024deepseek,shi2024eagle,jiang2023clip} or token compression~\cite{vasu2025fastvlm,wen2025stop,yang2025visionzip}, but fail to mitigate degradation in subsequent layers. Similarly, endpoint supervision~\cite{wang2024reconstructive} is insufficient for preventing intermediate information loss. While recent approaches align intermediate layers with VFMs~\cite{yoon2025visual}, they tend to bias representations toward VFM semantics rather than preserving input fidelity. This underscores the need for mechanisms that leverage external VFM knowledge to explicitly maintain the integrity of the original visual representation.

\section{Method}

\subsection{Pipeline Overview}
We introduce Spatial-Spectral Visual Anchor Learning (SSVAL), a framework designed to mitigate visual information degradation. As illustrated in Fig.~\ref{fig:main}, the core of our framework is Visual Anchor Prompt Injection (VAPI). During training, prompts act as knowledge vessels that absorb rich representations from VFMs via backpropagation, thereby enhancing the initial visual representations. During inference, these prompts function as visual anchors that effectively mitigate visual information degradation across layers. To provide additional vision-specific supervision during training, we further incorporate Spatial and Spectral Representation Alignment (SpaRA and SpeRA) as auxiliary training losses at intermediate LLM layers, addressing the limitation that standard text-based training leaves visual representations without explicit visual supervision.

\subsection{Visual Anchor Prompt Injection}
\begin{figure*}[t]
  \centering
  \includegraphics[width=\linewidth]{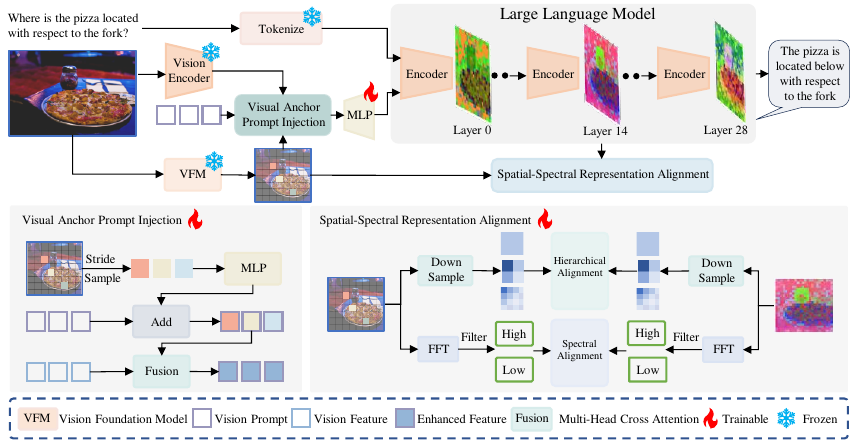}
  \caption{Overview of SSVAL framework. Visual prompts are combined with stride-sampled VFM features to inject visual prior knowledge, then fused with vision encoder outputs and fed into the LLM through the projector. To provide complementary visual supervision during training, we incorporate auxiliary spatial and frequency-domain alignment at intermediate LLM layers.}
  \Description{Detailed pipeline diagram of the SSVAL framework showing visual prompts combined with VFM features through a projection network, fused with vision encoder outputs via cross-attention, and fed into the LLM with auxiliary spatial and frequency-domain alignment losses at intermediate layers.}
  \label{fig:main}
\end{figure*}

The proposed Visual Anchor Prompt Injection module (see Fig.~\ref{fig:main}) consists of three core components: Prompts, a VFM feature sampling and projection module, and a Prompt Fusion Module. We denote $N_p$ as the number of prompts, $N_v$ as the number of visual tokens, $D$ as the dimension of prompts and visual features, $D_v$ as the VFM feature dimension, $D_h$ as the hidden dimension of the projection network, $H$ as the number of attention heads, and $\mathbf{Z}_v \in \mathbb{R}^{N_v \times D}$ as the visual features extracted by the vision encoder.

We introduce a set of visual prompts with dimension $[N_p, D]$, initialized randomly with a standard deviation of 0.02. These prompts are continuously optimized through backpropagation during training, ultimately serving as vessels for vision knowledge.

During training, we extract vision features from a pretrained VFM. Specifically, the VFM outputs features with dimension $[N_v, D_v]$, from which we uniformly sample with stride $N_v / N_p$ to obtain a feature subset of dimension $[N_p, D_v]$. Subsequently, a two-layer projection network ( $D_v \rightarrow D_h \rightarrow D$) maps these features to the same dimensional space as the prompts. Finally, the projected VFM features are added to the prompts to form the combined prompts.

The combined prompts incorporate both prompts and VFM residual information; during inference, we only use the prompts without relying on the VFM. Crucially, the prompts serve as visual anchors that provide stable reference points throughout the LLM layers, effectively mitigating visual representation deviation during inference.

To effectively inject the knowledge contained in prompts into vision features, we design a fusion mechanism based on multi-head cross attention. Vision features $\mathbf{Z}_v$ serve as queries, while visual prompts $\mathbf{p} \in \mathbb{R}^{N_p \times D}$ serve as keys and values, interacting through an $H$-head cross attention mechanism. The final output adopts a gated residual connection:
\begin{equation}
\mathbf{Z}_v' = \mathbf{Z}_v + \text{Gate}(\mathbf{Z}_v) \odot \text{CrossAttn}(\mathbf{Z}_v, \mathbf{p}),
\end{equation}
where the gating mechanism allows the model to adaptively control the degree of prompt information fusion.

\subsection{Spatial-Spectral Representation Alignment}
While VAPI addresses representation deviation through the anchor mechanism, standard text-based training still lacks explicit vision-specific supervision at intermediate layers. To provide complementary visual supervision, we introduce Spatial-Spectral Representation Alignment (SSRA), consisting of Spatial Representation Alignment (SpaRA) and Spectral Representation Alignment (SpeRA).

SpaRA aligns intermediate layer representations between the student (LLM) and teacher (VFM) from a multiscale spatial perspective. We denote $\mathbf{s}_i$ and $\mathbf{t}_i$ as the student and teacher features for the $i$-th token. We perform multiscale downsampling with pooling factors of $1\times 1$, $2\times 2$, and $4\times 4$ to capture semantic information at different granularities. Each scale uses an independent projection network to map student features into the teacher's feature space. For feature-level alignment, we maximize the cosine similarity between normalized student and teacher features. For relation-level alignment, we sample spatial neighbor pairs $(i, j)$ and constrain their pairwise similarities to match those in the teacher:
\begin{equation}
\begin{aligned}
\mathcal{L}_{\text{feat}} &= -\sum_{i} \cos(\hat{\mathbf{s}}_i, \hat{\mathbf{t}}_i), \\
\mathcal{L}_{\text{rel}} &= \text{MSE}(\cos(\mathbf{s}_i, \mathbf{s}_j), \cos(\mathbf{t}_i, \mathbf{t}_j)),
\end{aligned}
\end{equation}
where $\hat{\mathbf{s}}_i = \text{normalize}(\mathbf{s}_i)$ and $\hat{\mathbf{t}}_i = \text{normalize}(\mathbf{t}_i)$, $w_k$ is the weight for scale $k$, and $\alpha$, $\beta$ are the coefficients for feature and relation-level alignment respectively. The total SpaRA loss is:
\begin{equation}
\mathcal{L}_{\text{SpaRA}} = \sum_{k \in \{1,2,4\}} w_k \cdot (\alpha \cdot \mathcal{L}_{\text{feat}}^k + \beta \cdot \mathcal{L}_{\text{rel}}^k).
\end{equation}

While SpaRA focuses on spatial alignment, SpeRA provides complementary supervision from the frequency domain. We denote $\mathbf{S}, \mathbf{T} \in \mathbb{R}^{\sqrt{N_v} \times \sqrt{N_v} \times D_v}$ as the reshaped 2D feature maps of student (after projection) and teacher respectively. We first apply LayerNorm to eliminate magnitude discrepancies, reshape the token sequence into a 2D feature map, and transform it via 2D FFT:
\begin{equation}
\mathbf{F}_X = \text{FFT2D}(\mathbf{X}), \quad X \in \{S, T\}.
\end{equation}

We decompose the spectrum into low-frequency (LF) and high-frequency (HF) components using frequency masks $\mathbf{M}_{\text{LF}}$, $\mathbf{M}_{\text{HF}}$ based on the normalized radius $r$ from the center:
\begin{equation}
\begin{aligned}
\mathbf{M}_{\text{LF}}(u,v) &= \mathbb{1}[r(u,v) \leq 0.5] ,\\
\mathbf{M}_{\text{HF}}(u,v) &= \mathbb{1}[r(u,v) > 0.5],
\end{aligned}
\end{equation}
where LF components correspond to the global structure and HF components correspond to edges and local details.

\begin{table*}[t]
\centering
\caption{Comparison with baseline methods on six multimodal benchmarks. SSVAL achieves optimal results across different vision encoders (CLIP and SigLIP)~\cite{radford2021learning,tschannen2025siglip} and LLM scales (Qwen2.5-3B and Qwen2.5-7B)~\cite{bai2025qwen2}, demonstrating significant improvements on nearly all benchmarks and validating the effectiveness and generalization capability of our method.}
\label{tab:main_result}
\resizebox{\textwidth}{!}{%
\begin{tabular}{c|cc|cccccc}
\toprule
\rowcolor[gray]{.9} Method & Vision Encoder & LLM & CV-Bench$^\text{2D}$ & MMVP & MMStar & MME & POPE & MM-Bench \\
\midrule
Baseline & \multirow{3}{*}{CLIP} & \multirow{3}{*}{Qwen2.5-3B} & 51.53\% & 28.00\% & 36.40\% & 1579.16 & \textbf{86.97\%} & 74.59\% \\
Viral & & & \underline{52.85\%}{\tiny\textcolor{teal}{(+1.32)}} & \underline{32.00\%}{\tiny\textcolor{teal}{(+4.00)}} & \underline{37.07\%}{\tiny\textcolor{teal}{(+0.67)}} & 1531.05{\tiny\textcolor{red}{(-48.11)}} & 84.97\%{\tiny\textcolor{red}{(-2.00)}} & \underline{75.38\%}{\tiny\textcolor{teal}{(+0.79)}} \\
SSVAL & & & \cellcolor{cyan!15}\textbf{54.80\%}{\tiny\textcolor{teal}{(+3.27)}} & \cellcolor{cyan!15}\textbf{33.33\%}{\tiny\textcolor{teal}{(+5.33)}} & \cellcolor{cyan!15}\textbf{38.93\%}{\tiny\textcolor{teal}{(+2.53)}} & \cellcolor{cyan!15}\textbf{1613.94}{\tiny\textcolor{teal}{(+34.78)}} & \cellcolor{cyan!15}\underline{86.60\%}{\tiny\textcolor{red}{(-0.37)}} & \cellcolor{cyan!15}\textbf{77.41\%}{\tiny\textcolor{teal}{(+2.82)}} \\
\midrule
Baseline & \multirow{3}{*}{CLIP} & \multirow{3}{*}{Qwen2.5-7B} & 58.97\% & 33.47\% & 39.20\% & 1743.56 & \underline{85.88\%} & \underline{78.54\%} \\
Viral & & & \underline{60.50\%}{\tiny\textcolor{teal}{(+1.53)}} & \underline{36.07\%}{\tiny\textcolor{teal}{(+2.60)}} & \underline{39.67\%}{\tiny\textcolor{teal}{(+0.47)}} & \underline{1765.65}{\tiny\textcolor{teal}{(+22.09)}} & 84.92\%{\tiny\textcolor{red}{(-0.96)}} & 78.54\%{\tiny(+0.00)} \\
SSVAL & & & \cellcolor{cyan!15}\textbf{65.44\%}{\tiny\textcolor{teal}{(+6.47)}} & \cellcolor{cyan!15}\textbf{41.33\%}{\tiny\textcolor{teal}{(+7.86)}} & \cellcolor{cyan!15}\textbf{40.93\%}{\tiny\textcolor{teal}{(+1.73)}} & \cellcolor{cyan!15}\textbf{1822.72}{\tiny\textcolor{teal}{(+79.16)}} & \cellcolor{cyan!15}\textbf{86.10\%}{\tiny\textcolor{teal}{(+0.22)}} & \cellcolor{cyan!15}\textbf{79.65\%}{\tiny\textcolor{teal}{(+1.11)}} \\
\midrule
Baseline & \multirow{3}{*}{SigLIP2} & \multirow{3}{*}{Qwen2.5-7B} & 69.82\% & \underline{52.67\%} & \underline{43.73\%} & \underline{1924.64} & \underline{88.53\%} & \underline{82.01\%} \\
Viral & & & \underline{70.17\%}{\tiny\textcolor{teal}{(+0.35)}} & 47.33\%{\tiny\textcolor{red}{(-5.34)}} & 43.13\%{\tiny\textcolor{red}{(-0.60)}} & 1856.21{\tiny\textcolor{red}{(-68.43)}} & 88.07\%{\tiny\textcolor{red}{(-0.46)}} & 81.77\%{\tiny\textcolor{red}{(-0.24)}} \\
SSVAL & & & \cellcolor{cyan!15}\textbf{71.14\%}{\tiny\textcolor{teal}{(+1.32)}} & \cellcolor{cyan!15}\textbf{55.33\%}{\tiny\textcolor{teal}{(+2.66)}} & \cellcolor{cyan!15}\textbf{44.87\%}{\tiny\textcolor{teal}{(+1.14)}} & \cellcolor{cyan!15}\textbf{1950.21}{\tiny\textcolor{teal}{(+25.57)}} & \cellcolor{cyan!15}\textbf{88.67\%}{\tiny\textcolor{teal}{(+0.14)}} & \cellcolor{cyan!15}\textbf{82.65\%}{\tiny\textcolor{teal}{(+0.64)}} \\
\bottomrule
\end{tabular}%
}
\end{table*}

The filtered spectra are transformed back to the spatial domain via inverse FFT:
\begin{equation}
\begin{aligned}
\mathbf{S}_f &= \text{IFFT2D}(\mathbf{F}_S \odot \mathbf{M}_f), \quad f \in \{\text{LF}, \text{HF}\}, \\
\mathbf{T}_f &= \text{IFFT2D}(\mathbf{F}_T \odot \mathbf{M}_f), \quad f \in \{\text{LF}, \text{HF}\}.
\end{aligned}
\end{equation}

The frequency domain losses are computed as:
\begin{equation}
\begin{aligned}
\mathcal{L}_{\text{LF}} &= \text{MSE}(\mathbf{S}_{\text{LF}}, \mathbf{T}_{\text{LF}}), \\
\mathcal{L}_{\text{HF}} &= \text{MSE}(\log(|\mathbf{S}_{\text{HF}}| + \epsilon), \log(|\mathbf{T}_{\text{HF}}| + \epsilon)),
\end{aligned}
\end{equation}
where logarithmic compression prevents HF noise from dominating, and $\epsilon$ is a small constant for numerical stability. The total SpeRA loss is:
\begin{equation}
\mathcal{L}_{\text{SpeRA}} = \gamma_{\text{LF}} \cdot \mathcal{L}_{\text{LF}} + \gamma_{\text{HF}} \cdot \mathcal{L}_{\text{HF}},
\end{equation}
where $\gamma_{\text{LF}}$ and $\gamma_{\text{HF}}$ are the weights for LF and HF losses.

\subsection{Training Loss}
The total training loss combines the standard next token prediction loss~\cite{alayrac2022flamingo,liu2023visual} with the alignment losses~\cite{yu2024representation}:
\begin{equation}
\mathcal{L}_{\text{total}} = \mathcal{L}_{\text{NTP}} + \lambda_{\text{SpaRA}} \cdot \mathcal{L}_{\text{SpaRA}} + \lambda_{\text{SpeRA}} \cdot \mathcal{L}_{\text{SpeRA}},
\end{equation}
where $\mathcal{L}_{\text{NTP}}$ is the cross-entropy loss for language modeling, and $\lambda_{\text{SpaRA}}$, $\lambda_{\text{SpeRA}}$ are the weight coefficients for the respective alignment losses.

Through joint optimization, the learned visual prompts serve as the primary mechanism for mitigating representation deviation: during training, they act as knowledge vessels that absorb rich representations from VFMs via the VAPI module, while the auxiliary SpaRA and SpeRA losses provide supplementary vision-specific supervision at intermediate layers; during inference, the learned prompts—now embedded with VFM knowledge—function as visual anchors that help preserve visual information as it propagates through the model.

\section{Experiments}

\subsection{Implementation Details}
\paragraph{Train Details.}
Our method follows the LLaVA~\cite{liu2023visual} paradigm to construct an MLLM. Specifically, we employ CLIP~\cite{jiang2023clip} as the vision encoder to extract image features and Qwen2.5~\cite{bai2025qwen2} as the MLLM backbone, which offers advantages over Vicuna used in the original LLaVA. The visual representation projector adopts a lightweight three-layer MLP with SiLU activation functions~\cite{liu2024improved}, mapping vision features into the language model's embedding space. Before formal training, we first pretrain the projector using the LLaVA Visual Instruct Pretrain LCS-558K dataset with a batch size of 256 to achieve preliminary modality alignment. We then adopt the visual instruction tuning paradigm using the LLaVA-665K dataset~\cite{liu2023visual}, which contains diverse visual question answering, image captioning, and reasoning tasks. To preserve pre-trained knowledge of the language model and improve training efficiency, we apply LoRA~\cite{hu2022lora} fine-tuning (rank=128, $\alpha$ = 256) to the LLM, updating only a small number of adapter parameters. Training uses a learning rate of 2e-4 with AdamW optimizer. We use 4 visual anchor prompts. All experiments are conducted on 8 NVIDIA H20 GPUs.

\paragraph{Vision Foundation Models.}
We employ a series of pretrained VFMs as sources for visual knowledge injection and supervision signals. Different VFMs adopt different patch sizes: DINOv2 and CLIP use a patch size of 14, while DINOv3 uses a patch size of 16. To match the 576 visual tokens produced by CLIP-ViT-L/14 at $336 \times 336$ resolution in the CLIP+Qwen2.5-7B configuration, we adopt the same $336 \times 336$ input resolution for models with patch size 14, and resize inputs to $384 \times 384$ for models with patch size 16, ensuring the same number of visual tokens are generated.

\paragraph{Loss Configuration.}
In the total loss function, the weight coefficients for SpaRA Loss and SpeRA Loss are set to $\lambda_{\text{SpaRA}} = 0.3$ and $\lambda_{\text{SpeRA}} = 0.15$, respectively. For SpaRA Loss, it consists of three spatial scale components corresponding to pooling factors of $1 \times 1$, $2 \times 2$, and $4 \times 4$, with equal weights of 1:1:1. For SpeRA Loss, it comprises low-frequency and high-frequency component losses, also with equal weights of 1:1.

\subsection{Datasets and Evaluation Metrics}
To comprehensively validate the effectiveness of SSVAL, particularly its ability to mitigate visual degradation and enhance fine-grained perception, we conduct experiments on six widely-used multimodal benchmarks categorized by focus. First, to assess fine-grained visual perception and spatial reasoning, we employ CV-Bench~\cite{tong2024cambrian} and MMVP~\cite{tong2024eyes}. Specifically for CV-Bench, we utilize the 2D subset as the 3D subset is less relevant to the visual information degradation problem studied in this work, while for MMVP, we adopt Pair Accuracy as the primary metric. Second, to evaluate the stability of visual representations and the model's robustness against object hallucination, we use the Popular subset of POPE~\cite{li2023evaluating}, which serves as a proxy for verifying if our visual anchors effectively prevent feature drift. Finally, to ensure our method improves comprehensive multimodal capabilities across diverse tasks, we include MMStar~\cite{chen2024we}, MME~\cite{fu2025mme}, and MM-Bench~\cite{liu2024mmbench}. For these comprehensive benchmarks, we report overall accuracy or total scores as the final results.

\begin{table}[t]
\centering
\caption{Ablation study on different numbers of visual anchor prompts. All experiments are conducted on CLIP+Qwen2.5-7B. The model achieves stable improvements across various prompt quantities.}
\label{tab:ablation}
\scriptsize
\resizebox{\linewidth}{!}{%
\begin{tabular}{c|cccc}
\toprule
\rowcolor[gray]{.9} Prompts & CV-Bench$^\text{2D}$ & MMVP & MMStar & MME \\
\midrule
Baseline & 58.97\% & 33.47\% & 39.20\% & 1743.56 \\
\midrule
\multicolumn{5}{l}{\textit{Ablation on Different Prompts}} \\
24 & 63.00\% & 39.20\% & 39.40\% & 1803.79 \\
20 & 61.06\% & 36.67\% & \textbf{41.07\%} & \textbf{1851.18} \\
16 & \underline{63.28\%} & \underline{41.33\%} & 40.80\% & 1783.17 \\
12 & 61.75\% & 38.00\% & \textbf{41.07\%} & \underline{1824.95} \\
8 & 61.54\% & \textbf{42.67\%} & 40.67\% & 1767.88 \\
\rowcolor{cyan!15} 4 & \textbf{65.44\%} & 41.33\% & 40.93\% & 1822.72 \\
\bottomrule
\end{tabular}%
}
\end{table}

\subsection{Main Results}
We compare three methods: (1) Baseline~\cite{liu2023visual}, standard visual instruction tuning with only text-based supervision; (2) VIRAL~\cite{yoon2025visual}, which introduces visual alignment losses at intermediate LLM layers; and (3) SSVAL (Ours), which uses visual prompts as anchors to absorb VFM knowledge while mitigating representation deviation.

\paragraph{Results on CLIP Vision Encoder.}
As shown in Table \ref{tab:main_result}, SSVAL achieves consistent improvements across both LLM scales. At the 3B scale, SSVAL outperforms Baseline and VIRAL on all six benchmarks, with notable gains on CV-Bench$^\text{2D}$ (+3.27\%) and MMStar (+2.53\%). While VIRAL improves on most metrics, it suffers a drop on MME ($-$48.11), revealing instability with limited LLM capacity. When scaling to Qwen2.5-7B, SSVAL's advantages become more pronounced, achieving +6.47\% on CV-Bench$^\text{2D}$ and +7.86\% on MMVP—benchmarks designed to evaluate fine-grained visual perception and spatial reasoning. This suggests that as the LLM grows deeper, representation deviation becomes more severe, making SSVAL's anchor mechanism increasingly effective.

\paragraph{Generalization to SigLIP2 Vision Encoder.}
We further evaluate SSVAL with SigLIP2~\cite{tschannen2025siglip}, a stronger vision encoder. SSVAL continues to achieve consistent gains across all benchmarks, demonstrating effectiveness even when initial visual representations are already strong. Notably, VIRAL shows performance drops on five out of six benchmarks (e.g., MMVP $-$5.34\%, MME $-$68.43). We attribute this to the limitation of direct alignment: when the vision encoder already produces high-quality representations, forcefully aligning intermediate features can introduce conflicting gradients that disrupt existing representations. In contrast, SSVAL's prompt-based anchor mechanism serves as complementary references that preserve visual information without overwriting the original structure.

\paragraph{Benchmark-Specific Analysis.}
SSVAL achieves the most significant improvements on CV-Bench$^\text{2D}$ and MMVP, which emphasize spatial reasoning and fine-grained visual matching—tasks that are particularly sensitive to visual information loss in deeper LLM layers. These consistent gains support our core hypothesis that visual anchor prompts effectively suppress representation deviation and preserve task-relevant spatial and semantic details throughout forward propagation. The stronger performance on these perception-oriented benchmarks further indicates that SSVAL is especially beneficial for retaining subtle object relationships and local visual cues. Meanwhile, the stable improvements on comprehensive benchmarks, including MMStar, MME, and MM-Bench, demonstrate that mitigating visual degradation not only enhances fine-grained perception but also contributes to broader multimodal reasoning and general visual-language understanding.

\begin{table}[t]
\centering
\caption{Ablation study on different VFMs for visual anchor prompt injection.}
\label{tab:ablation_vfm}
\scriptsize
\resizebox{\linewidth}{!}{%
\begin{tabular}{c|cccc}
\toprule
\rowcolor[gray]{.9} VFM & CV-Bench$^\text{2D}$ & MMVP & MMStar & MME \\
\midrule
CLIP-L & 59.46\% & 37.33\% & 38.87\% & 1782.93 \\
DINOv3-L & 61.54\% & \underline{40.07\%} & \underline{39.67\%} & \underline{1784.00} \\
DINOv2-L & \underline{62.31\%} & \textbf{46.00\%} & \textbf{42.53\%} & \textbf{1875.22} \\
\rowcolor{cyan!15} DINOv2-B & \textbf{65.44\%} & 41.33\% & 40.93\% & 1822.72 \\
\bottomrule
\end{tabular}
}
\end{table}

\subsection{Ablation Study}

\begin{table}[t]
\centering
\caption{Ablation study on different components. VAPI: Visual Anchor Prompt Injection; SpaRA: Spatial Representation Alignment; SpeRA: Spectral Representation Alignment. VAPI provides the dominant improvement, with auxiliary alignment losses offering complementary gains.}
\label{tab:component}
\resizebox{\columnwidth}{!}{%
\begin{tabular}{l|cccc}
\toprule
\rowcolor[gray]{.9} Method & CV-Bench$^\text{2D}$ & MMVP & MMStar & MME \\
\midrule
Baseline & 58.97\% & 33.47\% & 39.20\% & 1743.56 \\
+VAPI & 62.10\% & \textbf{45.33\%} & 40.93\% & 1805.64 \\
+SpaRA & 57.44\% & 32.00\% & 38.13\% & 1828.21\% \\
+SpeRA & 59.04\% & 31.33\% & 35.67\% & 1823.08\% \\
+SpaRA+SpeRA & 60.22\% & 35.33\% & 39.20\% & 1779.56\% \\
+VAPI+SpaRA & \underline{63.35\%} & \underline{41.33\%} & \textbf{41.60\%} & \underline{1820.17} \\
+VAPI+SpeRA & 47.71\% & 9.33\%  & 31.60\% & 1489.98\% \\
\rowcolor{cyan!15} +VAPI+SpaRA+SpeRA & \textbf{65.44\%} & \underline{41.33\%} & \underline{40.93\%} & \textbf{1822.72} \\
\bottomrule
\end{tabular}%
}
\end{table}

\begin{table}[t]
\centering
\caption{Ablation study on multi-scale configurations in Spatial Representation Alignment (SpaRA). Scale numbers indicate the downsampling factors applied to visual features. The combination [1, 2, 4] achieves the best performance by capturing both fine-grained details and global structures.}
\label{tab:multiscale}
\footnotesize
\resizebox{\linewidth}{!}{%
\begin{tabular}{c|cccc}
\toprule
\rowcolor[gray]{.9} Scale & CV-Bench$^\text{2D}$ & MMVP & MMStar & MME \\
\midrule
{[}1{]} & 63.14\% & 38.67\% & 40.40\% & \textbf{1857.96} \\
{[}1, 2{]} & 62.73\% & 36.00\% & 39.73\% & 1821.85 \\
{[}1, 4{]} & \underline{64.39\%} & \underline{40.00\%} & \underline{40.07\%} & 1816.03 \\
\rowcolor{cyan!15} {[}1, 2, 4{]} & \textbf{65.44\%}  & \textbf{41.33\%}  & \textbf{40.93\%}  & \underline{1822.72}  \\
\bottomrule
\end{tabular}%
}
\end{table}

\paragraph{Prompt Quantity.}
To assess the effect of prompt quantity, we vary the number of prompts from 4 to 24. As shown in Table \ref{tab:ablation}, four prompts achieve the best CV-Bench$^\text{2D}$ result (65.44\%), while eight prompts perform best on MMVP (42.67\%). Larger prompt sets can benefit broader benchmarks by providing richer visual references. Overall, performance remains stable across different settings, demonstrating the robustness of VAPI. Considering both accuracy and efficiency, we use four prompts by default.

\paragraph{Vision Foundation Model.}
To evaluate knowledge transfer from different VFMs, we use several models as alignment targets. As shown in Table \ref{tab:ablation_vfm}, CLIP-L performs worst on most metrics, suggesting that homogeneous alignment provides limited complementary knowledge. External VFMs yield stronger results: DINOv2-B offers the best overall balance and is used by default, while DINOv2-L achieves the highest scores on MMVP, MMStar, and MME. DINOv3-L performs comparably to DINOv2-B, demonstrating that visual anchor prompts can benefit from diverse VFM architectures.

\paragraph{Component Analysis.}
To validate the contribution of each component, we conduct incremental ablation experiments. As shown in Table \ref{tab:component}, VAPI is the primary contributor: relative to the Baseline, it improves CV-Bench$^\text{2D}$ by 3.13 points, MMVP by 11.86 points, MMStar by 1.73 points, and MME by 62.08 points. These gains support our hypothesis that learnable visual anchors can mitigate visual information degradation. The alignment losses have complementary but benchmark-dependent effects. SpaRA combined with VAPI improves CV-Bench$^\text{2D}$ from 62.10\% to 63.35\% and MMStar from 40.93\% to 41.60\%, while the full VAPI+SpaRA+SpeRA configuration achieves the best CV-Bench$^\text{2D}$ score (65.44\%) and a strong MME score (1822.72). In contrast, MMVP does not improve beyond VAPI alone, and adding SpeRA without SpaRA causes a substantial drop, highlighting the importance of the interaction between the alignment objectives.

\paragraph{Multi-scale Configuration.}
To investigate the impact of multiscale configurations in SpaRA, we conduct ablation experiments with different scale combinations. As shown in Table \ref{tab:multiscale}, single-scale alignment [1] achieves reasonable performance but cannot capture visual information at different granularities. Interestingly, adding an adjacent scale [1, 2] does not improve performance and even degrades MMVP (36.00\% vs 38.67\%), suggesting that similar scales may introduce redundant supervision. In contrast, combining scales with larger intervals [1, 4] yields better results, indicating that complementary information across granularities is more beneficial. The full multiscale configuration [1, 2, 4] achieves the best overall performance across CV-Bench, MMVP, and MMStar, demonstrating that multiscale supervision captures both fine-grained details and global structures.

\begin{table}[t]
\centering
\caption{Ablation study on frequency components in Spectral Representation Alignment (SpeRA). Low-frequency alignment provides the primary contribution by capturing global semantic structures, while high-frequency alone degrades performance due to noise sensitivity. Combining both achieves the best results through complementary supervision.}
\label{tab:freq_component}
\scriptsize
\resizebox{\linewidth}{!}{%
\begin{tabular}{c|cccc}
\toprule
\rowcolor[gray]{.9} Frequency & CV-Bench$^\text{2D}$ & MMVP & MMStar & MME \\
\midrule
Baseline & 58.97\% & 33.47\% & 39.20\% & 1743.56 \\
\midrule
Low-frequency & \underline{62.31\%} & \underline{35.33\%} & 38.87\% & \underline{1802.68} \\
High-frequency & 57.02\% & 26.00\% & \underline{39.47\%} & 1691.41 \\
\rowcolor{cyan!15} LF + HF (Full SpeRA) & \textbf{65.44\%} & \textbf{41.33\%} & \textbf{40.93\%} & \textbf{1822.72} \\
\bottomrule
\end{tabular}%
}
\end{table}

\paragraph{Frequency Component Analysis.}
To understand the individual contributions of low-frequency (LF) and high-frequency (HF) components in SpeRA, we conduct ablation experiments using each frequency band independently. As shown in Table \ref{tab:freq_component}, low-frequency alignment alone yields consistent improvements over baseline in CV-Bench $^\text{2D}$ (+3.34\%) and MME (+59.12), confirming that global semantic structures captured by the LF component provide effective supervisory signals for visual representation alignment. In contrast, high-frequency alignment alone leads to significant performance degradation on most benchmarks, particularly on MMVP ($-$7.47\%) and MME ($-$52.15), indicating that high-frequency components are more susceptible to noise and can disrupt visual representations when used without the stabilizing effect of low-frequency supervision. This observation also validates our design choice of applying logarithmic compression to high-frequency losses, which mitigates the adverse impact of high-frequency noise. When combining both frequency bands, the full SpeRA achieves the best overall performance, demonstrating that LF and HF components provide complementary supervision: LF captures global semantic coherence while HF, when properly constrained, supplements fine-grained edge and texture details.

\subsection{Efficiency Analysis}
In SSVAL, SpaRA and SpeRA function as training-only alignment losses and introduce \textit{zero inference overhead}. The external VFM is also discarded after training. At inference time, only VAPI’s prompts and the lightweight gated cross-attention module are retained.

\begin{table}[t]
\centering
\caption{Efficiency analysis. SSVAL introduces only 5.25M (+1.55\%) additional inference parameters.}
\label{tab:efficiency}
\scriptsize
\resizebox{\linewidth}{!}{%
\begin{tabular}{l|ccc}
\toprule
\rowcolor[gray]{.9} Method & Train Params & Infer Params & Infer Overhead \\
\midrule
Baseline & 339.48M & 339.48M & -- \\
VIRAL & 352.60M & 339.48M & +0 params \\
SSVAL & 399.29M & 344.74M & +5.25M (+1.55\%) \\
\bottomrule
\end{tabular}%
}
\end{table}

\begin{table}[t]
\centering
\caption{Inference FLOPs comparison. SSVAL adds only 3.65 GFLOPs (+0.04\%) extra FLOPs at inference.}
\label{tab:flops}
\scriptsize
\resizebox{\linewidth}{!}{%
\begin{tabular}{l|ccc}
\toprule
\rowcolor[gray]{.9} Method & Inference FLOPs & Extra FLOPs & Overhead \\
\midrule
Baseline & 8.34 TFLOPs & -- & -- \\
VIRAL & 8.34 TFLOPs & 8.26 MFLOPs & +0.0001\% \\
SSVAL & 8.35 TFLOPs & 3.65 GFLOPs & +0.04\% \\
\bottomrule
\end{tabular}%
}
\end{table}

As shown in Tables \ref{tab:efficiency} and \ref{tab:flops}, SSVAL’s inference overhead is negligible: only +5.25M parameters (+1.55\%) and +3.65 GFLOPs (+0.04\%). This marginal cost yields substantial performance gains (e.g., +6.47\% on CV-Bench$^\text{2D}$, +7.86\% on MMVP), demonstrating a highly favorable efficiency-performance trade-off.
In contrast, VIRAL has no inference overhead but lacks mechanisms to mitigate representation deviation during inference. This reflects SSVAL’s design: SpaRA and SpeRA enhance representations during training without extra cost, while VAPI adds minimal overhead to continuously anchor visual information during inference, effectively preventing degradation and yielding notable performance gains (e.g., +6.47\% on CV-Bench$^\text{2D}$ and +7.86\% on MMVP).

\begin{figure*}[t]
\centering
\includegraphics[width=\linewidth]{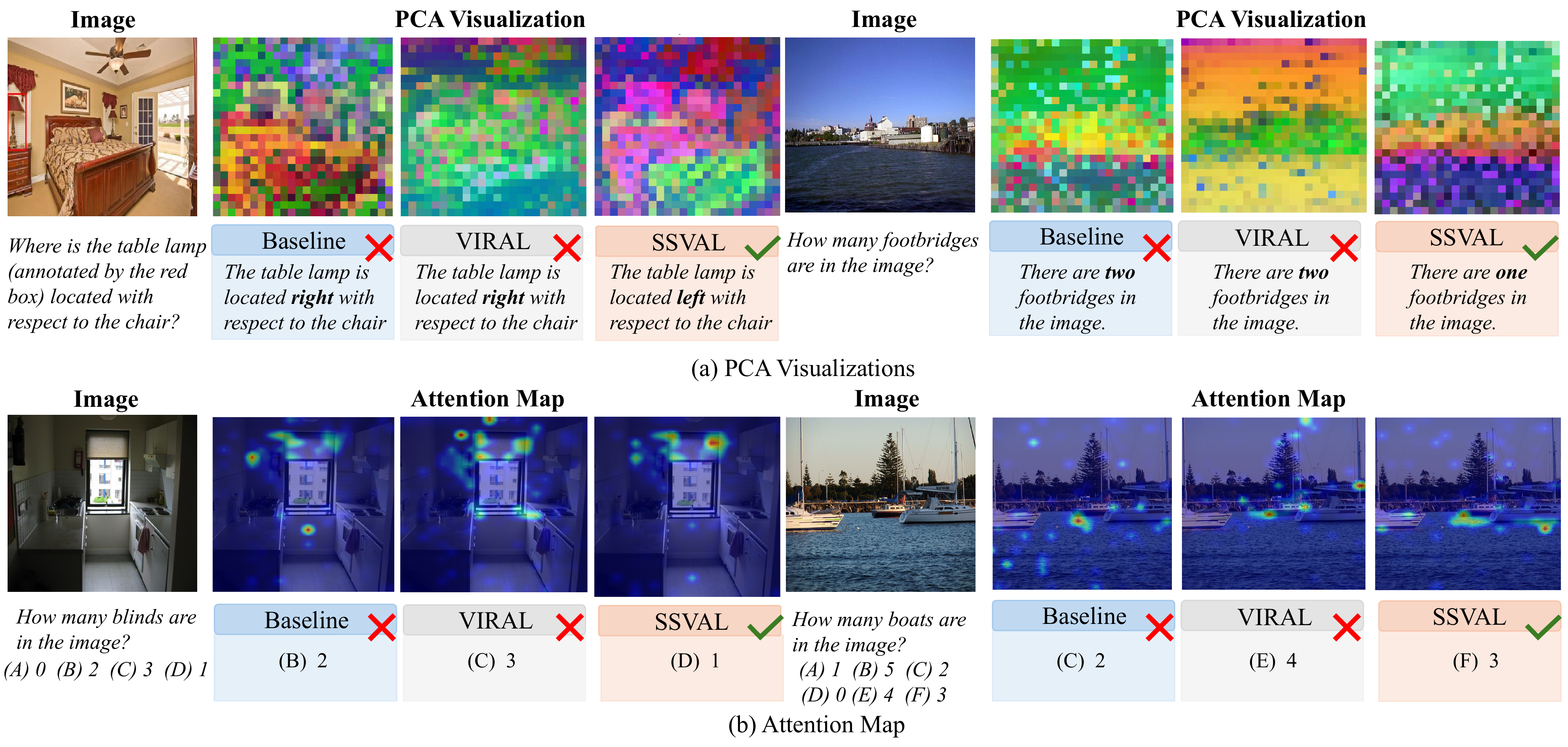}
\caption{Qualitative comparison among Baseline, VIRAL, and SSVAL(Ours). (a) PCA Visualizations reveal that SSVAL maintains robust semantic and spatial structures in intermediate layers, preventing the visual information decay observed in other methods. (b) Attention Maps show that SSVAL achieves superior object localization with sharply focused attention weights on relevant regions, resulting in correct answers where baselines fail.}
\Description{Side-by-side qualitative comparison showing PCA visualizations of visual token embeddings and attention maps for Baseline, VIRAL, and SSVAL. SSVAL produces more structured PCA clusters and sharply focused attention on relevant image regions compared to the diffuse patterns of other methods.}
\label{fig:pca}
\end{figure*}

\begin{figure}[t]
\centering
\includegraphics[width=1.0\columnwidth]{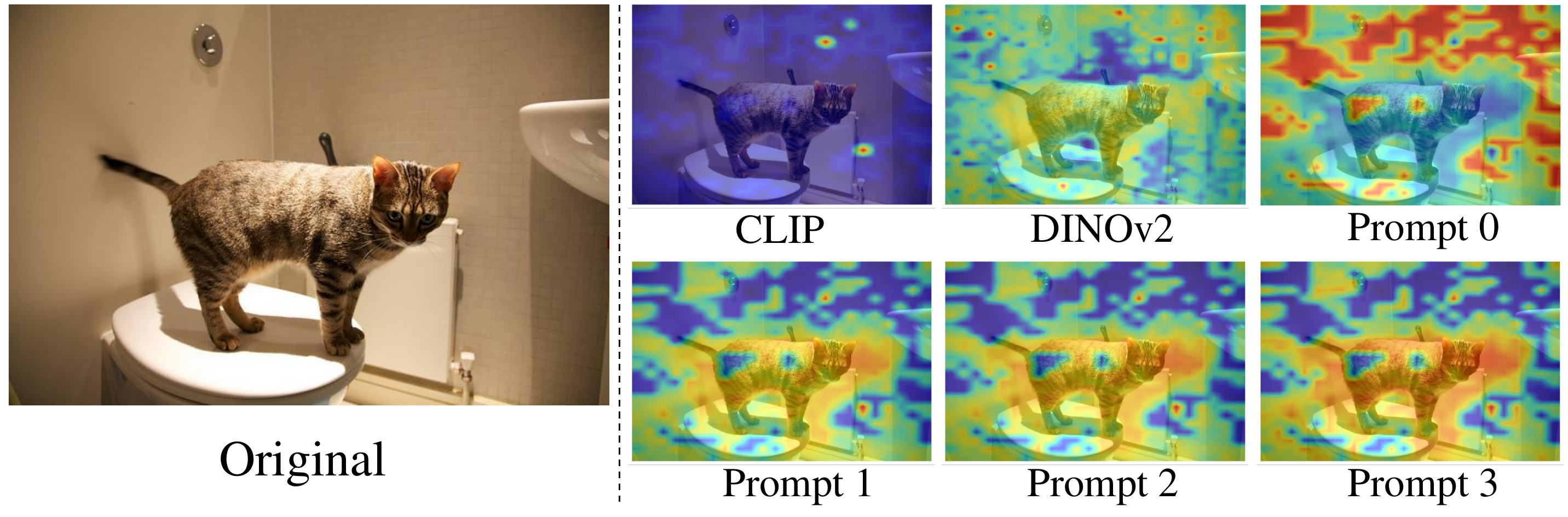}
\caption{Attention map visualization showing prompt-guided visual attention. Prompt 0 learns to suppress background regions, while Prompts 1-3 focus on foreground objects, inheriting DINOv2's objectness awareness.}
\Description{Grid of attention heatmaps for four learnable prompts overlaid on input images. Prompt 0 highlights background regions for suppression, while Prompts 1 through 3 concentrate on foreground objects, demonstrating complementary roles inherited from DINOv2.}
\label{fig:prompt}
\end{figure}

\subsection{Qualitative Results}
Through qualitative analysis, we further validate the effectiveness of SSVAL. As shown in Figure \ref{fig:pca}(a), SSVAL successfully solves challenging problems that both VIRAL and Baseline fail to address, particularly in scenarios requiring precise spatial localization and fine-grained visual understanding. PCA visualizations of final-layer visual tokens show that SSVAL produces embeddings with clearer clustering boundaries, stronger structural organization, and greater semantic coherence, indicating more effective preservation of critical visual information. This suggests that the proposed visual anchors help maintain discriminative representations as visual features propagate through deeper LLM layers.

Figure \ref{fig:pca}(b) further shows that visual anchor prompts help the model maintain focused attention on relevant regions during inference, whereas Baseline produces more scattered and unstable attention patterns. By consistently emphasizing task-relevant objects, SSVAL improves both localization accuracy and visual reasoning reliability. These observations are consistent with the quantitative findings and further confirm SSVAL’s improved visual understanding capabilities.

\subsection{Prompt Attention Analysis.}
To gain deeper insights into the working mechanism of prompts, we visualize the attention patterns of individual prompts. As shown in Figure \ref{fig:prompt}, learnable prompts serve as an auxiliary mechanism built upon CLIP's visual features, helping CLIP significantly alleviate its inherent deficiencies in spatial perception by absorbing foreground-background separation capabilities from DINOv2. The original CLIP model exhibits diffuse attention patterns that spill into background regions, while DINOv2 demonstrates strong objectness awareness with clear foreground-background separation. Through VAPI, prompts develop complementary roles: Prompt 0 acts as a background suppression mask, while Prompts 1-3 inherit DINOv2's foreground objectness. Combined with CLIP's semantic features, this achieves precise subject localization comparable to DINOv2. This analysis demonstrates that visual anchor prompts not only mitigate representation deviation but also enhance the model's spatial perception by bridging the gap between different VFM capabilities.

\section{Conclusion}

In this paper, we reveal that direct VFM alignment improves what representations encode but not how stably they are preserved across LLM layers. To address this limitation, we propose SSVAL, a unified framework with a clear architectural division: \textbf{VAPI} serves as the core inference-time module, where prompts absorb complementary VFM knowledge through gated cross-attention during training and act as persistent visual anchors during inference; \textbf{SpaRA} and \textbf{SpeRA} serve as training-time auxiliary losses that provide multi-scale spatial and frequency-domain supervision at intermediate LLM layers, further strengthening visual representations without introducing additional inference cost. This design effectively separates knowledge absorption from inference efficiency while maintaining a lightweight inference pipeline. Extensive experiments demonstrate that SSVAL consistently outperforms existing methods across multiple benchmarks, vision encoders, and LLM scales with negligible overhead. Qualitative analyses further show that the learned prompts develop specialized and complementary roles, helping integrate diverse VFM capabilities and preserve fine-grained visual information throughout the network.

\begin{acks}
This work was supported by Great Wall Motor.
\end{acks}

\bibliographystyle{ACM-Reference-Format}
\bibliography{reference}
\clearpage
\appendix
\section{Additional Implementation Details}

\subsection{Vision Encoder}
We employ different vision encoders to demonstrate the generalizability of our method. Both CLIP and SigLIP2 adopt a patch size of 14, but differ in input resolution: CLIP uses $336 \times 336$, while SigLIP2 uses $384 \times 384$. When using SigLIP2 as the vision encoder in SpaRA and SpeRA, the token counts between the student (SigLIP2) and teacher (DINOv2) are misaligned. To address this, we apply bilinear interpolation for spatial alignment:
\begin{equation}
\mathbf{F}_{\text{teacher}}^{\text{aligned}} = \text{Interpolate}(\mathbf{F}_{\text{teacher}}, \text{size}=(H_s, W_s))
\end{equation}
where $\mathbf{F}_{\text{teacher}} \in \mathbb{R}^{H_t \times W_t \times D}$ denotes the teacher (DINOv2) feature map, and $(H_s, W_s)$ represents the spatial dimensions of the student (SigLIP2) feature map. The bilinear interpolation resizes the teacher features to match the student's spatial resolution.
Additionally, for cases where removing the CLS token results in a non-perfect square number (e.g., $728 = 729 - 1$), we apply zero-padding to restore a complete spatial grid, enabling subsequent 2D convolution and FFT operations.

\subsection{Algorithm}
We provide the pseudocode for the SSVAL training pipeline in Algorithm~\ref{alg:ssval}.

\begin{algorithm}[hbtp]
\caption{SSVAL Training Pipeline}
\label{alg:ssval}
\footnotesize
\begin{algorithmic}[1]
\REQUIRE Images $\mathbf{I}$, input tokens, labels
\ENSURE Total loss $\mathcal{L}_{\text{total}}$
\STATE $\mathbf{Z}_v \leftarrow \text{VisionEncoder}(\mathbf{I})$; $\mathbf{P} \leftarrow \text{VisualAnchorPrompts}$
\IF{training}
    \STATE $\mathbf{P} \leftarrow \mathbf{P} + \text{Proj}(\text{Sample}(\text{VFMEncoder}(\mathbf{I}), N_p))$
\ENDIF
\STATE $\mathbf{Z}_v' \leftarrow \text{VAPI}(\mathbf{Z}_v, \mathbf{P})$
\STATE $\mathbf{H}_{\text{all}} \leftarrow \text{LLM}(\text{Embed}(\text{input}, \text{MMProjector}(\mathbf{Z}_v')))$
\STATE $\mathcal{L}_{\text{NTP}} \leftarrow \text{NTP}(\mathbf{H}_{\text{all}}, \text{labels})$
\IF{training}
    \STATE $\mathbf{H}_{\text{mid}} \leftarrow \mathbf{H}_{\text{all}}^{[\text{mid}]}[\text{visual tokens}]$
    \STATE $\mathcal{L}_{\text{SSRA}} \leftarrow \lambda_{\text{SpaRA}}\mathcal{L}_{\text{SpaRA}} + \lambda_{\text{SpeRA}}\mathcal{L}_{\text{SpeRA}}$
    \STATE \textbf{return} $\mathcal{L}_{\text{NTP}} + \mathcal{L}_{\text{SSRA}}$
\ENDIF
\STATE \textbf{return} $\mathcal{L}_{\text{NTP}}$
\end{algorithmic}
\end{algorithm}

\subsection{SpaRA, SpeRA, and Anchor Configuration}
For SpaRA, we sample 200 neighboring token pairs on the $24 \times 24$ visual-token grid. The relative offsets are uniformly selected from $[-2,+2]$ along both spatial dimensions, and the cosine similarity of each pair is aligned with the corresponding similarity computed from the frozen VFM teacher. This fixed-size sampling provides efficient spatial supervision while covering local relationships at multiple distances.

SpeRA operates on the same visual-token features after reshaping them into a 2D grid. We apply a two-dimensional FFT, separate low- and high-frequency components using fixed frequency masks, and compute the alignment loss after applying the inverse FFT. The masks and sampling budget are kept fixed throughout training.

Unless otherwise stated, we use post-normalization outputs from layers 16 and 28 of Qwen2.5-7B, select only the visual-token positions for representation alignment, and employ four learnable visual anchor prompts. These settings provide a balance between alignment coverage and computational efficiency.

\section{ADDITIONAL PROMPT ANALYSIS}
\begin{figure*}[t]
\centering
\includegraphics[width=\linewidth]{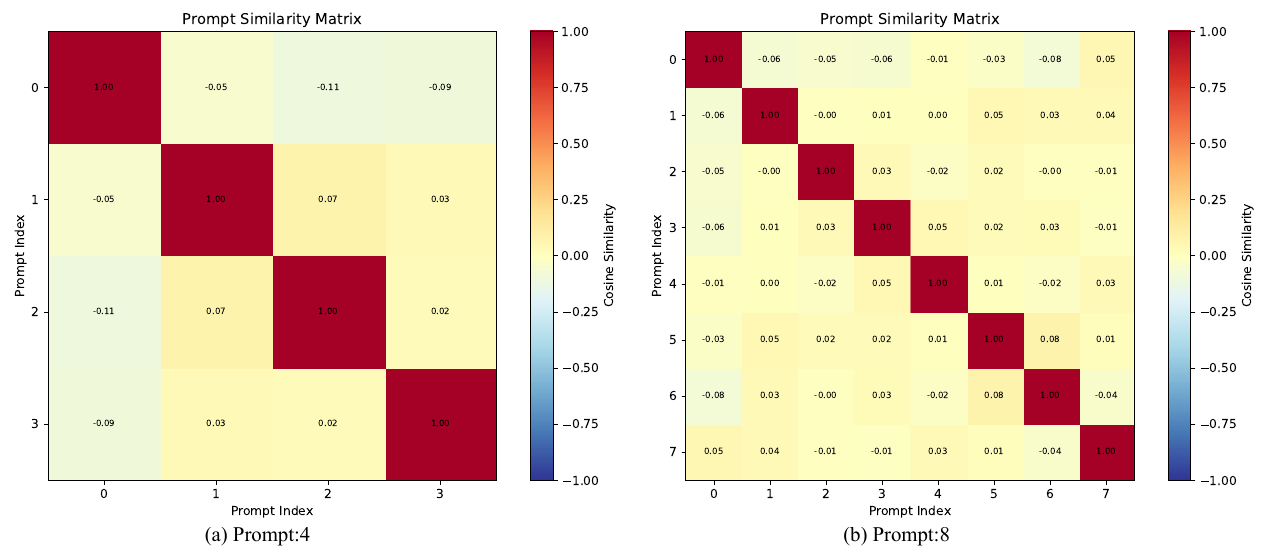}
\caption{Cosine similarity matrices of learned visual anchor prompts. (a) 4 prompts. (b) 8 prompts. The near-zero off-diagonal elements indicate that the learned prompts are approximately orthogonal, demonstrating that VAPI encourages each prompt to capture distinct and complementary visual semantics rather than redundant information.}
\Description{Two cosine similarity heatmap matrices for 4 and 8 learned prompts respectively. Diagonal elements are close to 1 while off-diagonal elements are near zero, indicating that the learned visual anchor prompts are approximately orthogonal and encode distinct visual semantics.}
\label{fig:pro_sim}
\end{figure*}

To verify that the learnable prompts capture diverse and complementary visual knowledge rather than redundant information, we visualize the cosine similarity matrix of the learned prompts. As shown in Fig.~\ref{fig:pro_sim}, the off-diagonal elements are consistently close to zero, indicating that the learned visual anchors are nearly orthogonal and encode distinct visual semantics.

\section{ADDITIONAL VISUALIZATIONS AND RESULTS}
\label{sec:additional_vis}

\subsection{Comparison with Stronger Baselines}
\begin{table*}[t]
\footnotesize
\centering
\caption{Comparison with baseline methods on six multimodal benchmarks. SSVAL achieves the strongest overall performance across six multimodal benchmarks, demonstrating effectiveness and generalization across vision encoders and Qwen2.5 scales.}
\vspace{-0.3cm}
\renewcommand{\arraystretch}{0.5}
\label{tab:main_result_1}
\setlength{\tabcolsep}{4pt}
\resizebox{\textwidth}{!}{%
\begin{tabular}{c|cc|cccccc}
\toprule
Method & Encoder & LLM & CV-Bench$^\text{2D}$ & MMVP & MMStar & MME & POPE & MM-Bench \\
Baseline & \multirow{3}{*}{CLIP} & \multirow{3}{*}{3B} & 51.53 & 28.00 & 36.40 & 1579.16 & \textbf{86.97} & 74.59 \\
Viral & & & 52.85 & 32.00 & \underline{37.07} & \underline{1531.05} & 84.97 & \underline{75.38} \\
ROSS & & & \underline{54.10} & 26.00 & 38.20 & 1676.61 & 86.23 & 76.58 \\
SSVAL & & & \textbf{54.80} & \textbf{33.33} & \textbf{38.93} & \textbf{1613.94} & \underline{86.60} & \textbf{77.41} \\
\midrule
Baseline & \multirow{3}{*}{CLIP} & \multirow{3}{*}{7B} & 58.97 & 33.47 & 39.20 & 1743.56 & 85.88 & 78.54 \\
Viral & & & 60.50 & 36.07 & \underline{39.67} & 1765.65 & 84.92 & 78.54 \\
ROSS & & & \underline{61.06} & \underline{37.33} & 38.93 & \underline{1814.81} & \textbf{86.67} & 78.40 \\
SSVAL & & & \textbf{65.44} & \textbf{41.33} & \textbf{40.93} & \textbf{1822.72} & \underline{86.10} & \textbf{79.65} \\
\midrule
Baseline & \multirow{3}{*}{SigLIP} & \multirow{3}{*}{7B} & 69.82 & 52.67 & 43.73 & 1924.64 & \underline{88.53} & 82.01 \\
Viral & & & \underline{70.17} & 47.33 & 43.13 & 1856.21 & 88.07 & 81.77 \\
ROSS & & & 70.01 & \underline{50.67} & \underline{44.07} & \underline{1940.97} & 87.90 & \textbf{83.34} \\
SSVAL & & & \textbf{71.14} & \textbf{55.33} & \textbf{44.87} & \textbf{1950.21} & \textbf{88.67} & \underline{82.65} \\
\bottomrule
\end{tabular}
}%
\end{table*}

To strengthen the comparison, we add ROSS, a recent and stronger baseline, to \textbf{Table~\ref{tab:main_result_1}}. SSVAL consistently outperforms Baseline, VIRAL, and ROSS on nearly all benchmarks. On the strongest CLIP+Qwen2.5-7B setting, SSVAL consistently outperforms the Baseline, VIRAL, and ROSS on key perception-heavy benchmarks. These results show that SSVAL's gains are not limited to a single alignment baseline, but generalize across stronger competitors, vision encoders, and LLM scales.

\subsection{Representation Visualization}
\begin{figure*}[t]
\centering
\includegraphics[width=\linewidth]{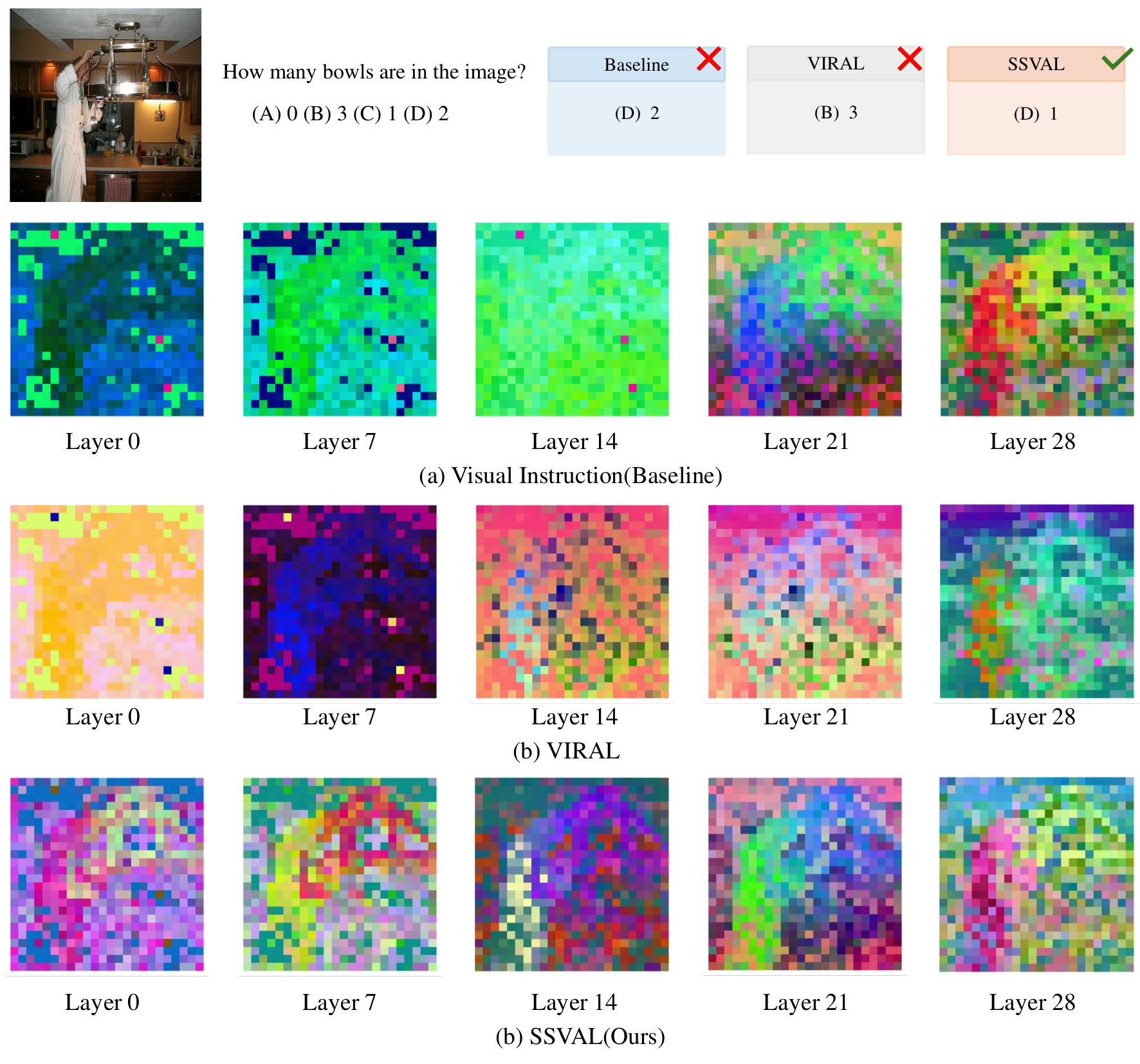}
\caption{Qualitative comparison among Baseline, VIRAL, and SSVAL(Ours).}
\Description{PCA visualization of visual representations across LLM layers from Layer 0 to Layer 28 for Baseline, VIRAL, and SSVAL. SSVAL maintains more semantically coherent and structured representations in deeper layers compared to the other two methods.}
\label{fig:pca_vis}
\end{figure*}

Fig.~\ref{fig:pca_vis} visualizes the evolution of visual representations from shallow to deep layers (spanning Layer 0 to Layer 28). In contrast to the standard Visual Instruction Tuning baseline (implemented with CLIP and Qwen2.5-7B) and VIRAL, our method maintains consistent superiority across the LLM's internal layers—particularly within deeper layers—generating representations that are significantly more semantically coherent and structured.

\section{LIMITATIONS}
Although SSVAL achieves significant performance improvements across multiple benchmarks, our method has the following limitations:

\noindent\textbf{Dependence on External VFM Quality.}
The performance of SSVAL depends partly on the teacher VFM quality: stronger VFMs yield better results, while weaker ones may provide less effective anchors. Future work could explore combining multiple complementary VFMs to improve anchor robustness.

\noindent\textbf{Additional Training Overhead.}
Although SSVAL does not require VFM during inference and introduces no additional computational overhead, the training phase still requires loading a frozen VFM to extract features for constructing visual anchors and computing alignment losses. This increases GPU memory usage and training time. Future work could explore offline pre-extraction of VFM features to reduce training overhead.

\noindent\textbf{Applicability to Mature MLLMs.}
Our method has only been validated during the visual instruction tuning stage. Whether SSVAL can still effectively mitigate visual representation deviation for mature MLLMs that have undergone multi-stage training remains to be explored. Future work could investigate how to apply the visual anchor mechanism to post-training fine-tuning or adaptation scenarios.

\end{document}